\documentclass[letterpaper, 10 pt, conference]{ieeeconf}  

\IEEEoverridecommandlockouts                              

\usepackage{hyperref}
\usepackage{graphicx}

\title{\LARGE \bf
Game Theory to Study Cooperation in Human-Robot Mixed Groups: \\ Exploring the Potential of the Public Good Game
}

\author{Giulia Pusceddu$^{1,3}$, Sara Mongile$^{2,3}$, Francesco Rea$^{2}$, and Alessandra Sciutti$^{2}$
\thanks{$^{1}$Robotics Brain and Cognitive Sciences Department (RBCS), Italian Institute of Technology, Genova, Italy}%
\thanks{$^{2}$Cognitive Architecture for Collaborative Technologies (CONTACT) Unit, Italian Institute of Technology, Genova, Italy}%
\thanks{$^{3}$Department of Computer Science, Bioengineering, Robotics and Systems Engineering
(DIBRIS), University of Genova, Genova, Italy}%
\thanks{AS and SM are supported by a Starting Grant from the European Research Council (ERC) under the European Union’s Horizon 2020 research and innovation programme. G.A. No 804388, wHiSPER.}
}

\begin{document}

\maketitle
\thispagestyle{empty}
\pagestyle{empty}

\begin{abstract}

In this study, we explore the potential of Game Theory as a means to investigate cooperation and trust in human-robot mixed groups. Particularly, we introduce the Public Good Game (PGG), a model highlighting the tension between individual self-interest and collective well-being. 
\\In this work, we present a modified version of the PGG, where three human participants engage in the game with the humanoid robot iCub
to assess whether various robot game strategies (e.g., always cooperate, always free ride, and tit-for-tat) can influence the participants' inclination to cooperate. We test our setup during a pilot study with nineteen participants.
A preliminary analysis indicates that participants prefer not to invest their money in the common pool, despite they perceive the robot as generous.
\\By conducting this research, we seek to gain valuable insights into the role that robots can play in promoting trust and cohesion during human-robot interactions within group contexts.
The results of this study may hold considerable potential for developing social robots capable of fostering trust and cooperation within mixed human-robot groups. 


\end{abstract}

\section{Introduction}

Game theory was initially designed to investigate firms' and consumers' behavior in the economic field \cite{von2007theory}. 
This theory provides a framework to examine how people make decisions when their actions have an effect on others, and, at the same time, their fates depend on the actions of other agents. 
Its nature makes it suitable for investigating social behavior in a controlled setting, 
with particular mention to games in which: 
(i) participants make decisions on their own, without the possibility of planning joint strategies ahead with the other players (non-cooperative), and 
(ii) there is the possibility that all the players get a positive score (non-zero-sum).
Non-zero-sum non-cooperative games,
such as Prisoner's Dilemma and Public Good Game (PGG),  
have been extensively used in social sciences to investigate cooperation and competition among humans \cite{sigmund2010social, dong2016dynamics, chaudhuri2002cooperation}.  

In the last years, even Human-Robot Interaction (HRI) researchers started employing Game Theory to evaluate cooperation and related phenomena (e.g., trust) between humans and artificial agents, mostly in dyadic interactions.
Schniter et al. \cite{schniter2020trust} found that, in an economic game, humans invest similarly in humans and robots.
Sandoval et al. \cite{sandoval2016reciprocity, sandoval2021robot}, using the Prisoner Dilemma and the Ultimatum Game paradigms, found that people tend to cooperate more with another person than with a robotic agent. However, apparently, they tend to be equivalently reciprocal with humans and robots. 
Similarly, Hsieh et al. \cite{hsieh2020human} found that people have a strong reciprocal tendency to social robots in economic games and this tendency might even surpass the influence of the reward value of their decisions.
\\
Correia et al. \cite{correia2019exploring} ran a version of the Public Good Game in which a human played with two EMYS robots. Researchers manipulated the score of the rounds.
One robotic agent always cooperated, while the other always defected.
It was found that the former was preferred in terms of social attributes with respect to the latter, regardless of the game result.

Even in this paper, we propose an HRI version of the PGG, involving the humanoid robot iCub and three human players. As better described in Section \ref{social-dyn-implications}, 
we aim to study the game strategy of the participants and their impression of the robot. 
In this preliminary phase of our research, we conducted a small pilot study to test the experimental design and collect participant feedback on the robot that behaved as a cooperator. Moreover, we plan on testing the cooperating and defecting robotic behaviors in a between study.  

\section{Public Good Game}

\subsection{Description of the game}
A minimum of three participants play the game.
Each player is given a certain amount of tokens, which they can contribute to a common pool or keep for themselves.
Each player independently decides how much of their resources they want to contribute to the common pool. The decision is made simultaneously and without communication between players.
Once all players have made their contribution decisions, the total amount of resources in the common pool is calculated by summing up the contributions of all players. This total is often multiplied by a factor greater than 1 to represent an increasing return on the collective investment.
The resources in the common pool are then distributed among all players, including those who contributed and those who did not. The total amount is divided equally among the players.
Each player's payoff is determined by the sum of their share from the common pool and any remaining resources they kept for themselves.

The game can be played as a single round or repeated over multiple rounds. In repeated games, players have the opportunity to observe the contributions of others in previous rounds and adjust their strategies accordingly.

\renewcommand{\thefootnote}{\roman{footnote}}
\subsection{Social dynamics implications}
\label{social-dyn-implications}
In the Public Good Game, players must consider the potential benefits of contributing to the common pool against the cost of losing resources to maximize their own payoffs.
This game highlights a cooperation dilemma. While it may be individually rational to contribute nothing to the common pool and free-ride on others' contributions (\emph{Nash equilibrium}\footnote{Nash equilibrium is a solution concept in game theory where no player can improve their outcome by unilaterally changing their strategy.}), if all players act in this manner, the common pool will be insufficient, and everyone's payoff will be lower.
It provides a framework for analyzing decision-making in situations where individuals must balance their self-interest with the common good. 

To reach the scenario where the highest payoff is achieved, all players must trust that the others will contribute with all of their tokens.
A one-shot version of this game could be employed to study trust based on pre-existing individual characteristics (e.g., personality traits, prior experiences, or attitudes towards robots) \cite{hancock2011meta}, while a multiple-round version could be helpful to investigate the evolution of trust during the game.  
We are particularly interested in understanding whether the trust toward a human-robot mixed group can be influenced by the robot's game strategy. Thus, we formulate the following research question:
\begin{center}
    Do the human players' contributions change according to the robot's game strategy? \\ 
\end{center} 
  More in particular, since, in a Public Good Game scenario with humans only Wu et al. \cite{wu2014role} found that players that display more cooperation lead to higher group average contribution, we ask:
  \begin{center}
     Does a robotic \emph{always cooperate} strategy promote other players' contributions?
\end{center} 
 
\section{Methodology}

\subsection{Participants}
We tested the setup with 19 participants (15 males, 3 females, 1 not declared) between 9 and 15 years old ($12.4 \pm 1.5$ y.o.) during a social event for students organized by  \href{https://www.orientamenti.regione.liguria.it/}{Orientamenti Regione Liguria} 

\subsection{Setup}
\label{methodology}
 \begin{figure}[t]
      \centering
      {\includegraphics[scale=0.60]{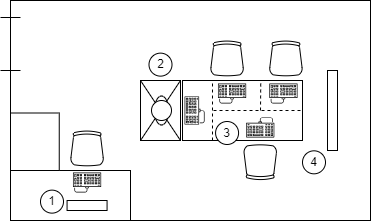}}
      \caption{A schema of the setup of the experiment. (1) experimenter seat; (2) iCub robot and its keyboard; (3) human players' positions with their keyboard; (4) display showing the GUI. 
      The dashed lines indicate the dividing cardboard. }
      \label{fig_interaction}
   \end{figure} 
In our version of the game, four players (iCub and three humans) play 10 rounds. 
They sit around a table positioned in front of a screen showing the Graphic User Interface (GUI). 
To avoid players peeking at the choices of others, we positioned dividing cardboard on the table.   
See Figure \ref{fig_interaction} for a schema of the setup.

\begin{figure}[t]
      \centering
      {\includegraphics[scale=0.30]{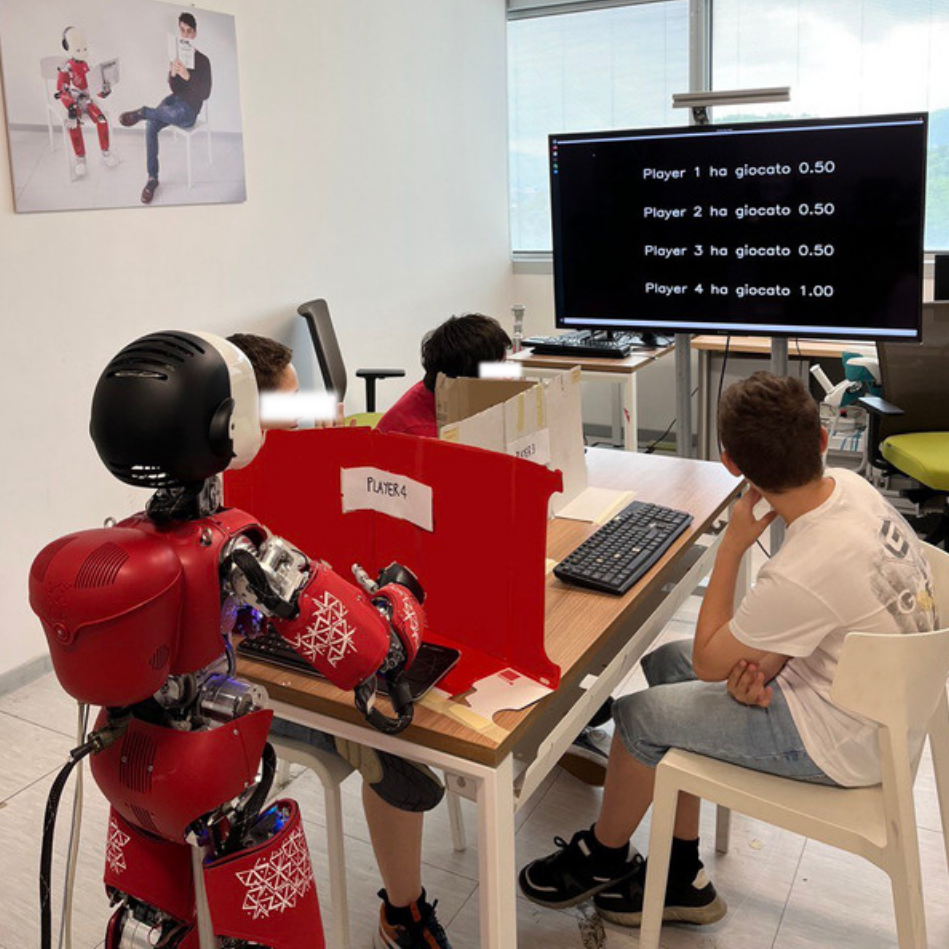}}
      \caption{A picture of the participants and the robot during the interaction.}
      \label{fig_picture}
   \end{figure}

The game's rules appear on the screen, accompanied by an explanation by the researchers. 
Participants play ten rounds, and each player - human and robotic -  has 1 € per round. They can decide whether to invest 0 €, 0.50 €, or 1 € in the common pool using a keyboard. 
After every player has made their choice, the value in the common pool is multiplied by a 1.6 factor. Next, the decisions and the scores are displayed on the screen so everyone can see other players' strategies and decide if to change theirs, an example is shown in figure \ref{fig_picture}.

\label{methodology}
At the end of the game, the final scores are shown on the screen. 

After the activity, participants answered a brief questionnaire. The questions included: (i) age and gender, (ii) if they had seen iCub before, (iii) how generous they thought the robot was on a 1-5 scale, and (iv) the perceived role of the robot, using the same item as \cite{alves2016role}. 

\subsection{Robot's behavior}
In this preliminary phase of our research, the behavior of the robot is pre-programmed.
The GUI and the robot's movements are controlled by a state machine coded in C++\footnote{The code is available upon request to the authors.}. 

During the game, iCub keeps an \emph{always cooperate} strategy, i.e., it always plays 1 € in the common pool.
It greets the other participants at the beginning and at the end of the activity. In the decision phase, it looks at its keyboard and pretends to press a key as the human players. 
To make iCub's behavior less repetitive, during this phase, it performs an action from a set of random behaviors (e.g., it says ``A second, please, let me think'', makes a focused facial expression, or clears its throat). 
After all the participants have made their decision, iCub raises its gaze, randomly looks at one of the human players, and then at the screen as if it is checking the scores. 

\section{Preliminary evaluations}
The setup worked as expected, and the participants demonstrated an understanding of the game.

\begin{figure}[h!]
      \centering
      {\includegraphics[scale=0.40]{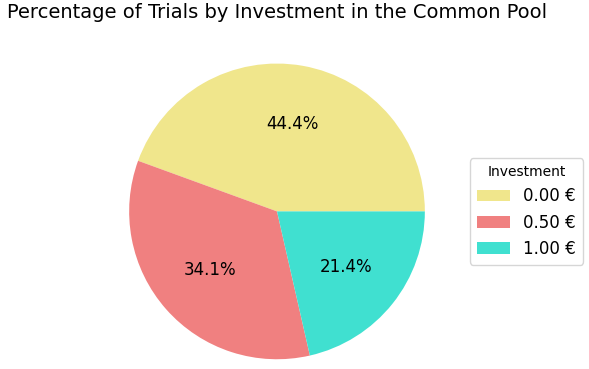}}
      \caption{This pie chart represents the investments in the common pool of all the human players. The three sectors represent the number of trials in percentage by the investment sum (0 €, 0.50 €, and 1 €). Since each one of the 19 participants played 10 rounds, the total number of trials is 190.}
      \label{pie}
   \end{figure}

We started analyzing the choices of the participants - 190 trials in total - and found that
players chose prevalently not to invest money in the common pool (44.4\%), they invested 0.50 € in 34.1\% of the cases, and played 1 € only in 21.4\% of the trials (Figure \ref{pie}). 

Results of the questionnaire showed that overall the players found iCub generous. To the question: ``How generous did iCub seem to you?'' they answered positively ($M = 4.2 , SD = 0.9$ on 5-point scale, where 5 meant ``very generous'' and 1 ``not generous at all''). 

\begin{figure}[h!]
      \centering
      {\includegraphics[scale=0.60]{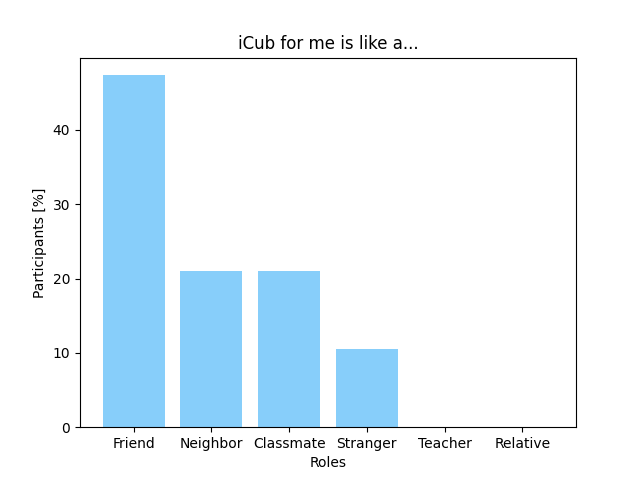}}
      \caption{This bar chart represents the role that the participants attributed to robot iCub after the game.}
      \label{bar}
   \end{figure}

Concerning the perceived role of the robot, 47.4\% of the participants has perceived iCub as a friend, the 21,1\% as a neighbor, the 21.1\% as a classmate, the 10.4\% as a stranger. Nobody perceived it as a teacher or a relative.
These results are summarized in Fig. \ref{bar}.

Despite they apparently perceived the robot as generous and most of them declared to have considered it as a friend, they decided not to invest their tokens in the common pool. 
As this work was a preliminary study, we do not have sufficient data to comprehensively explain this behavior; we plan to investigate this aspect in our future study.

\section{Future directions}

This pilot study allowed us to collect initial feedback on our game design. For instance, the experiment could be improved by adding some more pre-programmed robot behaviors; we believe this would make the interaction experience more engaging and natural.
Additionally, we plan on improving the GUI. The one we used for testing the setup is a plain, black-and-white template, although we think adding colors and sounds would make the interface more attention-drawing and easier to consult for participants. 

The preliminary data analysis revealed that participants perceived the robot as generous but were reluctant to invest their tokens in the common pool. Understanding the reasons for this behavior requires a bigger sample and further analysis, but these initial findings suggest that the robot's generosity alone may not be sufficient to encourage cooperative behavior in human-robot groups. To deeper investigate this aspect and answer the research questions presented in Section \ref{social-dyn-implications}, we plan on conducting a between study. The samples will differ by the robot's game strategy: (i) \emph{always cooperate}, (ii) \emph{always free ride} (i.e., it always plays nothing in the common pool), (iii) \emph{copycat} or \emph{tit for tat} (i.e., it copies the last move of the other players). 

In the pilot described in this paper, we tested our setup with children for a set of circumstances and opportunity; however, our intent is to investigate adults' group dynamics. If the research gives promising results, we will extend the study to a younger population. 

Overall, the collected preliminary feedback let us believe that our setup is valuable for investigating cooperation in HRI. The outcomes of this research might help to define how a robot should behave to promote trust and cohesion in human-robot mixed groups.






\bibliographystyle{IEEEtran}

\end{document}